\documentclass{edm_article}
\usepackage{cite}
\usepackage{amsmath,amssymb,amsfonts}
\usepackage{algorithm,algorithmic}
\usepackage{graphicx}
\usepackage{subcaption}
\usepackage{textcomp}
\usepackage[dvipsnames]{xcolor}
\usepackage{filecontents}
\usepackage{url}
\usepackage{colortbl}
\usepackage{makecell}
\usepackage{pgfplots}
\usepackage{multirow}
\usepackage{caption,latexsym}
\usepackage{wrapfig}
\usepackage{tikz}

\newcommand{\eat}[1]{}
\newtheorem{definition}{Definition}

\begin{document}
    
\title{Scalable and Equitable Math Problem Solving Strategy Prediction in Big Educational Data}

\numberofauthors{3}
\author{
\alignauthor Anup Shakya\\
    \affaddr{University of Memphis}\\
    \email{ashakya@memphis.edu}\\
\alignauthor Vasile Rus\\
    \affaddr{University of Memphis}\\
    \email{vrus@memphis.edu}\\
\alignauthor Deepak Venugopal\\
    \affaddr{University of Memphis}\\
    \email{dvngopal@memphis.edu}
}

\maketitle
\begin{abstract}
Understanding a student's problem-solving strategy can have a significant impact on effective math learning using Intelligent Tutoring Systems (ITSs) and Adaptive Instructional Systems (AISs). For instance, the ITS/AIS can better personalize itself to correct specific misconceptions that are indicated by incorrect strategies, specific problems can be designed to improve strategies and frustration can be minimized by adapting to a student's natural way of thinking rather than trying to fit a standard strategy for all. While it may be possible for human experts to identify strategies manually in classroom settings with sufficient student interaction, it is not possible to scale this up to big data. Therefore, we leverage advances in Machine Learning and AI methods to perform scalable strategy prediction that is also fair to students at all skill levels. Specifically, we develop an embedding called MVec where we learn a representation based on the mastery of students. We then cluster these embeddings with a non-parametric clustering method where we progressively learn clusters such that we group together instances that have approximately symmetrical strategies. The strategy prediction model is trained on instances sampled from these clusters. This ensures that we train the model over diverse strategies and also that strategies from a particular group do not bias the DNN model, thus allowing it to optimize its parameters over all groups. Using real world large-scale student interaction datasets from MATHia, we implement our approach using transformers and Node2Vec for learning the mastery embeddings and LSTMs for predicting strategies. We show that our approach can scale up to achieve high accuracy by training on a small sample of a large dataset and also has predictive equality, i.e., it can predict strategies equally well for learners at diverse skill levels.
\end{abstract}

\eat{
\begin{abstract}
    Predicting the strategy that the student is likely to use to solve a problem helps Adaptive Instruction Systems (AIS) to better adapt themselves to different types of learners based on their learning abilities. This can lead to a more dynamic, engaging and personalized experience for students. In this paper, we propose a framework for predicting student strategy as a sequence of knowledge components based on LSTM architecture. The problem-solving strategies have a high-degree of symmetry and specially in the context of middle-school students, there are likely to be large groups who follow similar strategies. The similarity in strategy is guided by the level of mastery of each student. And since mastery is a latent variable, we build a representation that integrates the symmetry between strategy and mastery. In particular, we develop a method similar to {\em Node2Vec}, which we call {\em MVec} to generate mastery-based embeddings. Further, we use non-parametric clustering with coarse-to-fine refinement to discover the symmetric groups and sample from those strategy invariant groups to obtain high quality training data. We use this to train a highly scalable LSTM-based model. We apply our model to real-world large-scale datasets from MATHia, a leading AIS math learning tool. The results show significant improvement in strategy prediction. Furthermore, this approach is fair to students at all levels of learning.
\end{abstract}
}
\keywords{Intelligent Tutoring Systems, Strategy Prediction, Equity, Representation Learning, Skill Mastery, Non-parametric Clustering, Fairness, Transformers, LSTM, Symmetry}

\section{Introduction}

The recent pandemic has spurred a remarkable growth in virtual learning and with it, the necessity to develop learning technologies that are effective even in the absence of face-to-face instruction. To this end, Intelligent Tutoring Systems (ITSs)~\cite{rus&al13} and more broadly Adaptive Instructional systems (AISs) will play a key role in education since they can scale up personalized instruction to large and diverse student populations. However, to adapt to a student, an AIS should be able to understand the student's thinking process which can be challenging. For instance, if we consider math learning, students can solve the same problem using several different approaches or {\em strategies}. Understanding these strategies can help an ITS/AIS adapt more effectively~\cite{ritter&al19}. For example, the type of strategy can reveal the expertise/knowledge of a student in a topic, incorrect strategies that indicate misconceptions can be corrected by the ITS, the student can be trained to change strategy based on the problem context, and students may be less frustrated if the ITS guides them towards strategies that are more naturally aligned to their thinking.



\begin{figure*}
    \includegraphics[scale=0.639]{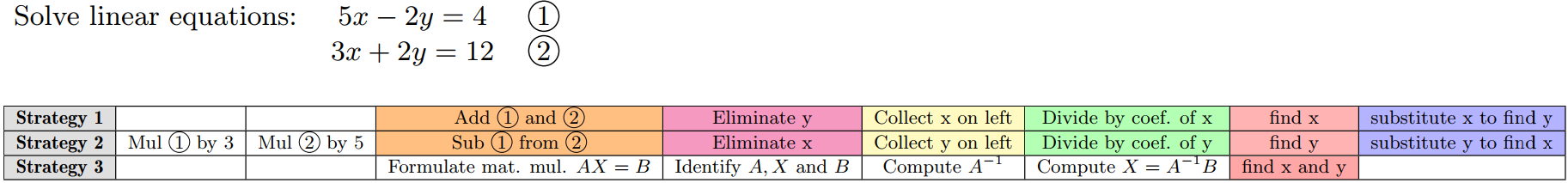}
    \caption{Illustrating symmetries in strategies where similar colors indicate similar steps. Strategies 1 and 2 are similar in that they use the elimination method but are not identical. Strategy 3 uses the matrix method which indicates a higher level of sophistication in student mastery.}
    \Description{Illustrating the similarity in different types of unidentical strategies.}
    \label{fig:strat-similarity}
\end{figure*}

In math problem solving, a {\em strategy} is a sequence of actions/steps that the student performs to solve a problem. An example of 3 different strategies is shown in Fig. \ref{fig:strat-similarity}. Human tutors can recognize different strategies followed by students and utilize these in one-on-one instruction. For instance, if a student is a visual learner, then they can teach the student to solve problems through visual aids, or if the student prefers an analytical approach to solve the same problem, then they can modify their teaching accordingly. However, adapting this approach for ITSs is challenging, particularly since identifying problem-solving strategies through computational methods is a complex problem. Specifically, there may be several strategies that are similar/symmetric without being completely identical. An example is illustrated in Fig.~\ref{fig:strat-similarity} to show similar and dissimilar strategies. As shown here, 2 of the 3 strategies are not exactly identical but implement the same idea and are thus symmetrical. The third strategy is quite different and asymmetrical to the first two strategies. Further, there may be several strategies that may not be conventional approaches to problem-solving but are indicative of unique ways in which students think about problems. Thus, if we identify a new strategy based on matching them with a set of previously known strategies, this approach may not be very effective when we want to scale up to big educational data. While there have been several approaches to detect strategies including using model tracing~\cite{corbett01} or sequence mining~\cite{wong&al19} methods, newer advances in deep neural networks (DNNs) can learn much more complex representations from large-scale data. Thus, leveraging such DNNs, we predict {\em novel} strategies more effectively.  

\eat{
\begin{figure*}
    \centering
    \scalebox{0.7}{
        \includegraphics{figs/strategy.pdf}
    }
    \caption{Example of a strategy used by a student to solve a math problem.}
    \label{fig:student-strategy}
\end{figure*}
}

Our goal in this paper is to develop a scalable and equitable model to predict strategies in math learning. Specifically, though DNN models are highly effective, they may tend to produce biased results. For instance, since most DNNs have a loss function that optimizes the overall loss, depending on the data distribution used during training, their results may be unfair to some sub-groups in the data. In our context, we want to avoid the model being unfairly biased where it can only identify strategies for certain student sub-groups. Specifically, we want to avoid {\em disparate mistreatment}~\cite{fairness2019} where the model accuracy is significantly different for different types of learners. In particular, learners may have disparity in their mastery or skill level which will influence their choice of strategy for a problem. For example, in Fig.~\ref{fig:strat-similarity}, the third strategy shown in the figure is more sophisticated than the other two and the student who applies this strategy is likely to have greater mastery in the topic. Therefore, we want to ensure that our model can predict strategies equally well for learners at all skill levels. To do this, we use a sampling approach, where instead of training the DNN over the full dataset (which may contain biases), we modify the underlying data distribution. Specifically, we sample the data such that sub-groups in the data are equally well-represented. Thus, when the DNN is trained over these samples instead of the full dataset, the DNN is forced to optimize its loss over all sub-groups. In general, sampling is a well-known approach used to scale up complex DNNs while training the model from large datasets~\cite{katharopoulos&al18}. Further, it has been shown that in some cases using too much data can lead to poor generalization~\cite{nakkiranKBYBS20}. In our case, a naive sampling approach where we sample students uniformly at random and train over strategies used by the sampled students will certainly be biased towards the skill level of the majority group and does not account for inequalities in skill levels.
Therefore, here, we develop an iterative non-parametric clustering method where we cluster the data into groups where each group corresponds to strategies corresponding to similar skills levels. Further, since strategies themselves are hard to compare exactly, we develop an approach where we use {\em approximate symmetries} to group strategies. We then train a DNN to predict a strategy by sampling from these diverse groups.



We implement our approach using the {\em DP-Means} Hierarchical Dirichlet Process framework~\cite{kulis&jordan12} to jointly cluster students and problems. Specifically, we project students (and problems) into an embedding space that we term {\em MVec} (Mastery Vectorization). To do this, we represent relationships between symbolic objects (students, problems, and concepts used in strategy) as a graphical structure. We then learn dense vectors using an embedding approach called {\em Node2Vec} \cite{grover&leskovec16} that assigns similar embeddings to nodes that have similar neighborhoods. We add mastery over concepts used in the strategy as weights in the graph estimated from a transformer model with attentions~\cite{vaswani&al17}. Thus, students with mastery over similar concepts in their strategies are assigned similar embeddings. 
We optimize the clusters incrementally where in each step, we adaptively change a penalty parameter based on the symmetries encoded by the clusters in the previous step. To quantify approximate symmetries, we develop a strategy alignment procedure with {\em positional encodings}~\cite{smith_waterman}. Once the clusters converge, we sample training instances from the clusters and train a Long Short Term Memory (LSTM) model that predicts strategies.

We evaluate our approach on two datasets from MATHia, a commercial AIS widely used for math learning in schools. The data is available through the PSLC datashop~\cite{StamperKBSLDYS11a}. The datasets are both large datasets that consist of millions of data instances (an instance is a student-problem pair and has multiple interactions in the dataset). Our results confirm that using our approach, we can sample a substantially smaller set of instances from the big dataset which we can use to train the strategy prediction model efficiently and achieve high accuracy in strategy prediction for students at diverse levels of mastery.


\section{Background}

\subsection{Related Work}
Ritter et al.~\cite{ritter&al19} provide a comprehensive survey on different approaches used to identify student strategies. Well-known approaches include the use of model tracing-based methods~\cite{corbett01} to identify strategies. In such cases, strategies may be pre-specified and the tutor can recognize correct and incorrect strategies. Model-tracing-based methods have also been adapted to recognize new strategies~\cite{ritter97}. Sequence learning approaches have been used in Open-Ended Learning Environments such as Betty's brain~\cite{leelawong&al08}. In \cite{wong&al19}, sequence pattern mining was applied to a MOOCs platform to analyze activity sequences of learners. For conversational tutors, natural language conversation interactions between tutors and students were mapped into a taxonomy of higher-level pedagogical concepts (e.g. scaffolding) by education experts~\cite{morrison&al15}. These concepts can also be seen as a form of strategy and models have been developed to predict these concepts from conversational tutors~\cite{rus&al17,maharjan&al18,venugopal&rus16}. Shakya et al.~\cite{shakya2021} developed an approach using importance sampling to sample data instances to scale up training of a strategy prediction model based on student interaction data from Mathia. Specifically, they formulated a Neuro-Symbolic AI model~\cite{VenugopalRS21} where symbolic formulas were used in conjunction with a DNN to train the model. However, unlike our approach \cite{shakya2021} has two fundamental limitations in identifying strategies. Particularly, their work does not use mastery to diversify the training samples which is important for equitable training. Further, it does not learn approximately symmetrical groups in a non-parametric manner. Thus, it cannot effectively group together symmetrical strategies which is necessary if we want to train the DNN from strategies that represent all such groups.

Mastery-based learning was proposed in the classic work by Bloom~\cite{Bloom68} to reduce achievement gaps between diverse students. The famous Bloom 2-sigma rule illustrates the benefits of such mastery-based learning. Ritter et al.~\cite{RitterYFB16} more recently provides a detailed insight into how mastery learning works in large-scale environments through their experiments on the MATHia platform. Knowledge tracing~\cite{corbett01} is a well-known approach for inferring the {\em knowledge state} of students over KCs which indicates the degree of mastery over the KCs. More recently, deep knowledge tracing~\cite{piech15} performed knowledge tracing using deep learning models. There is also a significant momentum in tackling the Knowledge Tracing problem in terms of graphs with the advent of GNNs \cite{nakagawa2019graph, liu2020improving}. In \cite{SONG2022108274}, node-level and graph-level GCNs have been used to learn exercise-to-exercise and concept-to-concept relational sub-graphs adding to the semantic value of the representations. The natural phenomenon of learning, forgetting and dynamic changes to a student's mastery of knowledge concepts is formulated using gating-controlled mechanisms in \cite{ZHAO2023103114}. Learning the pre-requisite structure of various associated skills has proven to be insightful to understand the problem-solving patterns~\cite{chen2015, penteado2016}. In \cite{PandeyK19}, an attention-based model was proposed to predict correct answers but this was not used to predict strategies which is the focus of our work.


Our approach to using symmetries to make deep learning more scalable is inspired by the Geometric Deep Learning (GDL)~\cite{bronstein21} framework. Specifically, GDL is a formal framework used to understand the effectiveness of DNNs from the perspective of symmetries. Here, we ground GDL in the context of improving the effectiveness of DNNs in strategy prediction from big, diverse data. More generally, being selective about training instances has been shown to improve scalability and generalization~\cite{nakkiranKBYBS20}. Deep importance sampling~\cite{katharopoulos&al18} has the same underlying principle as our approach  in that they propose to sample data to scale up training. However, unlike our approach they do not use symmetries as a basis for efficiently and equitably training the model. More recently, there has been work on improving fairness in DNNs by adaptively selecting batches during training to improve fairness measures such as minimizing gender disparity~\cite{Roh0WS21}. In principle, our approach also tries to achieve a similar goal in the context of educational data which is more challenging given that both mastery and strategies are complex variables.

\subsection{Overview of Embedding Models} 

We use the well-known embedding model Node2Vec~\cite{grover&leskovec16} to learn our mastery-based embedding MVec. Node2Vec is an embedding model for graphs and learns embeddings/dense vectors for nodes in the graph base on local neighborhoods. It is well-known to be a highly scalable approach for learning embeddings from large graphs. Node2Vec assigns similar vector representations for nodes with similar neighborhoods. Internally, it uses a skip-gram model called Word2Vec~\cite{mikolov&al13} to learn these representations. Word2Vec, which was originally developed for word embeddings, is used to predict neighboring nodes (also called context) from a given node. An autoencoder architecture is used in Word2Vec and the hidden layer learns the embedding. When neighborhoods are similar for two nodes, since their contexts are similar, the embedding learned for the two nodes will also be similar. Thus, Word2Vec projects the nodes into a continuous embedding space where similar/symmetrical nodes lie close to each other in the space.


\subsection{DP-Means}

DP-Means~\cite{kulisJ12} is a non-parametric clustering algorithm that does not require us to specify of the number of clusters. The DP-Means Hard Gaussian Processes (HDP) clustering learns a 2-step hierarchy where {\em local clusters} for multiple datasets are learned at the lower level and these clusters are associated with {\em global clusters} at the higher level. Let $x_{ij}$ denote the $i$-th instance of dataset $j$. The specific objective function of HDP is as follows.
\begin{equation}
    \label{eq:dpobj}
    \sum_{p=1}^g\sum_{x_{ij}\in\ell_p}||x_{ij}-\mu_p||_2^2+\lambda_{\ell}k+\lambda_{g}g
\end{equation}
where $\ell_p$ is the $p$-th global cluster, $k$ is the total number of local clusters, $\mu_p$ is the center of the $p$-th global cluster, $g$ is the total number of global clusters, $\lambda_{\ell}$ is a local penalty that controls the formation of local clusters and $\lambda_{g}$ is a global penalty that controls the formation of global clusters.

We can minimize the objective in Eq.~\eqref{eq:dpobj} HDP clustering as follows. For each $x_{ij}$, we compute the distance to the current global cluster means. If the minimal distance exceeds $\lambda_{\ell}+\lambda_g$, we create a new local cluster for $x_{ij}$ and a new global cluster $\ell_g$ associating it with the newly created local cluster. If the minimal distance is smaller than the sum of penalties, then we find the closest global cluster for $x_{ij}$, say $\ell_{g'}$. We then add $x_{ij}$ to a local cluster that is already a part of $\ell_{g'}$. If no such local clusters exist, we create a new one for $x_{ij}$ and associate it with $\ell_{g'}$. We then process the local clusters as follows. Let $c$ denote a local cluster. We compute the global cluster whose mean is at a minimal distance, $d'$ from $c$. Let the sum of distances of the points in the local cluster $c$ to its cluster center be $m$. If $d'$ is greater than the sum of the global cluster penalty and $m$, we create a new global cluster and assign $c$ to this new global cluster. This algorithm converges to a locally optimal solution for Eq.~\eqref{eq:dpobj} as shown in \cite{kulisJ12}.

\subsection{Positional Encodings}

Positional encodings~\cite{vaswani&al17} are used to encode positional information in a sequence using a continuous vector space. Specifically, using sine and cosine functions that alternate with frequencies, we can represent positions in a sequence as follows. Let the position of the $t$-th item in the sequence be encoded by the $d$ dimensional vector $\vec{p}_t$. The $k$-th dimension in $\vec{p}_t$ is computed as follows. If $k$ is even, the value is equal to the sinusoidal function $sin(\omega_k.t)$ and if $k$ is odd, the value is equal to the cosine function $cos(\omega_k.t)$, where $\omega_k$ $=$ $1/10000^{2k/d}$. The frequencies of the sine and cosine functions increase as $k$ increases. Positional encodings are widely used to augment the latent representation learned by a deep network with positional information for sequence learning.
\section{Proposed Approach}

Since strategy is a generic term, we define it more precisely. Specifically, we consider strategies in the context of structured interaction between students and tutors. In this case, a student interacts with a tutor and solves a problem by sequentially solving the steps that lead to the final solution. Thus, we can think of a strategy as a sequence of actions the student takes among possible sequences in an action-space. Operationally, each step in the sequence is associated with a specific {\em knowledge component} (KC)~\cite{koedinger&al12} which is defined by domain experts and corresponds to the concept/knowledge required to solve that step. Thus, in our discussion, a strategy corresponds to a sequence of KCs. Further, note that a step can be associated with multiple KCs in which case, we can just unroll the step to ensure that each step has a single KC. While it is possible to adapt our approach to perform structure prediction where instead of a single KC, a step can be mapped to a more complex structure (e.g. a graph), we leave this for future work and focus on the case where a single KC is mapped to a step in the strategy.

In this paper, the task that we want to solve is the following. Given a student $s$ and a problem $p$, we predict the sequence of KCs that $s$ will use to solve $p$. In particular, we assume that we have a large dataset $\mathcal{D}$ where we refer to an {\em instance} in the dataset as a pair $(s,p)\in \mathcal{D}$. We want to sample instances from $\mathcal{D}$ to train a model that takes as input $(s,p)\in \mathcal{D}$ and predicts strategies, i.e., variable-length sequences of KCs. We also assume that $\mathcal{D}$ contains correctness associated with each step in the strategies. Specifically, for an input $(s,p)$, for each step that $s$ takes to solve $p$, we know if $s$ was successful in solving that step correctly. We use this information to determine the mastery of a student and based on this, we develop an embedding (vector representation) for students and problems. We then jointly cluster the embeddings using a non-parametric approach such that instances where the strategies are approximately symmetric are clustered together. Finally, we train an LSTM model to predict strategies by sampling the clusters.
In the subsequent subsections, we first describe our embedding called MVec. Next, we apply DP-Means HDP clustering~\cite{kulisJ12} to the embeddings while also incorporating approximate symmetries in strategies.

\subsection{MVec Embeddings}

\begin{figure}
    \centering
    \includegraphics[scale=0.4]{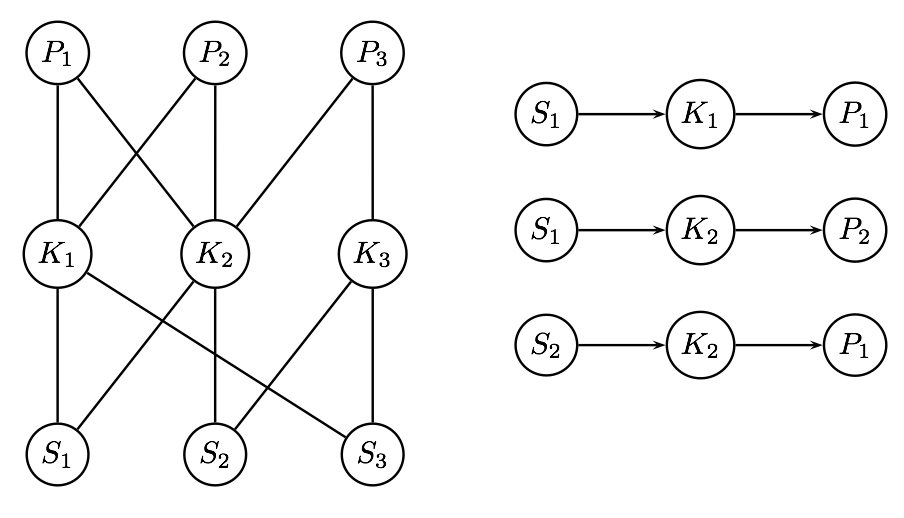}
    \caption{Illustrating a graph network of three students, problems, and KCs. The figure on the right shows some of the sampled random walks/paths.}
    \Description{Illustrating a graph network of three students, problems, and KCs.}
    \label{fig:graphex}
\end{figure}

To learn the MVec embedding, we use an approach that is similar to Node2Vec~\cite{grover&leskovec16}. Specifically, we construct a relational graph $\mathcal{G}$ $=$ ({\bf V},{\bf E}) as follows. Each student, problem, and KC in the training data is represented as a node $V\in{\bf V}$. For every student $S$ who uses KC $K$ as a step to solve problem $P$, there exist 2 edges $E, E'$ $\in$ {\bf E}, where $E$ connects the node representing the student to the node representing the KC and $E'$ connects the node representing the KC to the node representing the problem. An example graph over 3 students, problems, and KCs is shown in Fig.~\ref{fig:graphex}. We now sample paths in the graph and learn embeddings for these paths using word embedding models (Word2Vec)~\cite{mikolov&al13}. Specifically, the objective function is as follows.

\begin{equation}
    \max_{f}\sum_{u\in {\bf V}}log P(N_Q(u)|f(u))
\end{equation}
where $f:u\rightarrow \mathbb{R}^d$ is the vector representation for nodes $u\in{\bf V}$, $N_Q(u)$ denotes the neighbors of $u$ sampled from a distribution $Q$. Similar to Node2Vec, we assume that there is a factorized model that gives us a conditional likelihood that is identical to the likelihood function used in Word2Vec.
\begin{equation}
    P(n_i|f(u))=\frac{\exp(f(n_i)\cdot f(u))}{\sum_{v\in{\bf V}}\exp(f(v)\cdot f(u))}
\end{equation}

where $n_i$ is a neighbor of $u$. The conditional likelihood is optimized by predicting neighbors of $u$ using $u$ as input in an autoencoder neural network. The hidden layer learns similar embeddings for nodes with symmetrical neighborhoods. To do this, we generate walks on $\mathcal{G}$ as shown for the example in Fig.~\ref{fig:graphex}, and in each walk, given a node, we predict neighboring nodes similar to predicting neighboring words in sentences. To generate these walks, a simple sampling strategy $Q$ is to randomly sample a neighbor for a node. However, in our case, it turns out that each neighbor may have different importance when it comes to determining symmetry. Specifically, if a student has achieved mastery in applying a KC to a problem, then the corresponding edges should be given greater importance when determining symmetry between nodes in $\mathcal{G}$. To do this, we train a Sequence-to-Sequence attention model~\cite{vaswani&al17} from which we estimate the sampling probabilities for edges in $\mathcal{G}$.

\begin{figure*}
    \centering
    \scalebox{1.15}{
        \includegraphics{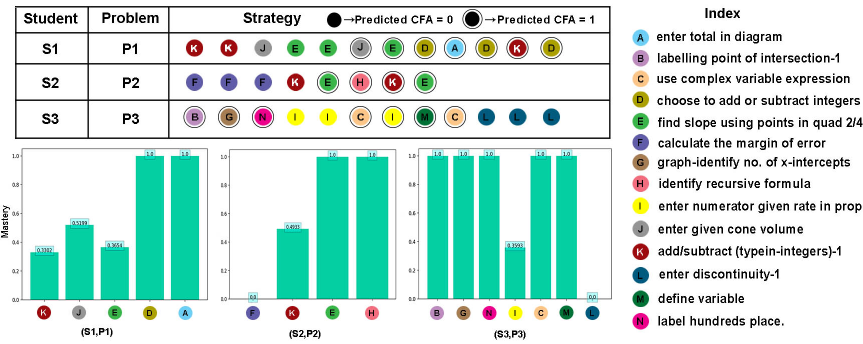} 
    }
    \caption{An example to illustrate the use of attention for mastery estimation. The bar charts show for each KC, the attention on a KC across steps that the student solves successfully (CFA=1) normalized by total attention for that KC. Larger values indicate that the model believes the student understands the KC as the attention on it is large when CFA=1 and vice versa.}
    \Description{Illustrating the use of attention for mastery estimation.} 
    \label{fig:latent-mastery}
\end{figure*}

The intuitive idea in quantifying mastery is illustrated in Fig.~\ref{fig:latent-mastery} which shows the opportunities given to 3 students to apply KCs in different problems. For each sequence of KCs, we predict if the student got the step correct or wrong on the first attempt (abbreviated as CFA for Correct First Attempt) when given an opportunity to apply the KC. The CFA values are performance indicators for the student, i.e., if they have mastered a KC, then they are likely to get the step correct in every opportunity they get to apply that KC. We train a model to predict the CFA values (CFA $=$ 1 indicates a correct application of the KC) given the KCs used in a problem. The predicted values from the model are shown for each KC. The bar graphs show mastery over the KCs. As seen here, the first student is inconsistent in applying the skill, {\em find the slope using points} (labeled as $E$) since the predictions for this oscillate between 0 and 1 whenever the student tries to apply this KC. On the other hand, student 2 consistently applies the same skill correctly and therefore the attention value is higher. We train the attention model from opportunities based on curriculum structure. Specifically, the curriculum consists of multiple units and each unit is further subdivided into sections.
For each student $S$, from every unit that the student has completed say $U$, we select a problem $P$ from each section that the student has worked on in $U$ and train the model to predict the CFA values for each KC used in $P$. 
We use the standard architecture described in ~\cite{vaswani&al17} for this model. Specifically, the input consists of the KC sequence, and the encoder maps this sequence to a latent representation and the decoder decodes the CFA values one at a time. The attention is given by
\begin{equation}
\label{eq:atteq}
    Attention( \gamma,\kappa,\eta) = softmax\left(\frac{\gamma \kappa^T}{\sqrt{d_k}}\right)\eta
\end{equation}
where $\gamma$, $\kappa$, and $\eta$ are the standard query, key, and value matrices respectively as defined in~\cite{vaswani&al17}, and $d_k$ is the dimensionality of the embedding that represents the latent representations. We use the encoder-decoder attention, i.e., the query is the decoder representation and the key is the encoder representation. The attention weights are an estimate of the alignment between encoded latent representations of mastery with the decoded representation of correctly applying a skill at each step in the problem. The projection of mastery over a KC $K$ based on the attention vectors is estimated by the following equation.
\begin{equation}
\label{eq:attn}
  \alpha(S,P,K) = \frac{\sum_i\sum_{v\in\pi(a_i)}v}{\sum_i\sum_{v\in\pi(a_i)}v+\sum_i\sum_{v'\in\bar{\pi}(a_i)}v'}  
\end{equation}
where $\pi(\cdot)$ extracts only those values in the input vector where the corresponding output for that step is predicted as 1, i.e., the model predicted that the student could solve the step correctly. $\bar{\pi}(\cdot)$ is the complement of $\pi(\cdot)$, i.e., it extracts attention values corresponding to steps that were predicted as mistakes made by the student and $i$ sums up all the instances where K is used.

We now sample paths from $\mathcal{G}$ using the factored distribution, i.e., $Q(S)*Q(K|S)*Q(P|K,S)$, where $Q(S)$ is the probability of sampling a student node, $Q(K|S)$ is the probability of sampling a KC $K$ given student $S$ and $Q(P|K,S)$ is the probability of sampling problem $P$ given $K,S$. We assume that $Q(S)$ is a uniform distribution over students. The conditional distributions are as follows.
\begin{equation}
\label{eq:q1}
Q(K|S)=1/n\sum_p\alpha(S,P,K)    
\end{equation}
\begin{equation}
\label{eq:q2}
Q(P|K,S)=\alpha(S,P,K)
\end{equation}
\eat{
$$
Q(x|y) = \begin{cases}
  1/n\sum_p\alpha(y,p,x)  &  \text{ y is a student node} \\
  1/m\sum_u\alpha(u,x,y) & \text{y is a KC node}
\end{cases}
$$
}
where $n$ is the number of opportunities given to student $S$ to apply KC $K$. The algorithm to generate MVec embeddings is shown in Algorithm \ref{alg:mvec}. As shown here, we sample a path in the graph as follows. We first sample student $S$ uniformly at random, then we sample a KC $K$ from $Q(K|S)$ and a problem from $Q(P|K,S)$. We then predict each node in the path using the neighboring nodes through a standard Word2Vec model. The resulting embeddings are learned in the hidden layer of the Word2Vec model. Note that for scalability, we do not construct/store the full graph $\mathcal{G}$ at any point. Instead, we only sample paths in an online manner as shown in Algorithm \ref{alg:mvec}.


\begin{algorithm}
    \caption{Generate MVec embeddings}
    \algsetup{
        linenodelimiter=.
    }
    \begin{algorithmic}[1]
        \renewcommand{\algorithmicrequire}{\textbf{Input:}}
        \renewcommand{\algorithmicensure}{\textbf{Output:}}
        \REQUIRE Relation Graph: $\mathcal{G} = (\bf V, \bf E)$ with student, problem and KCs as nodes, Embedding dimension: $d$, pre-trained attention-model $\mathcal{A}$
        \ENSURE Embeddings for each node $v \in \mathbb{R}^d$
        \\ \textit{Initialize}: set of walks, $\mathcal{W} = empty$
        \FORALL{$t$ $=$ 1 to $T$}
        \STATE Sample  a path $<S,K,P>$ in $\mathcal{G}$ from $Q(S)*Q(K|S)*Q(P|K,S)$ using Eq. (6) and (7).
        \STATE $\mathcal{W}$ $=$ $\mathcal{W}$ $\cup$ $<S,K,P>$
        \ENDFOR
        \STATE$v_e = word2vec(\mathcal{W}, d)$
        \RETURN $v_e$
    \end{algorithmic}
    \label{alg:mvec}
\end{algorithm}

\subsection{Non-Parametric Clustering}



We cluster the student and problem MVec embeddings jointly through a non-parametric approach based on symmetries defined as follows. For the dataset denoted by $\mathcal{D}$, let ${\bf S}$, ${\bf P}$ denote the set of students and problems respectively in $\mathcal{D}$.
\begin{definition}
A {\em strategy-invariant} partitioning w.r.t $\mathcal{D}$ is a partitioning $\{{\bf S}_i\}_{i=1}^{k_1}$ and $\{{\bf P}_j\}_{j=1}^{k_2}$ such that $\forall i,j$, if $S,S'\in{\bf S}_i$, $P,P'\in{\bf P}_j$, $S,S'$ follow equivalent strategies for $P,P'$ respectively.
\end{definition}
where $k_1$ and $k_2$ are the number of partitions/clusters for students and problems respectively.
The benefit of strategy-invariant partitioning is that we can scale up without sacrificing accuracy by training a prediction model only on samples drawn from the partitions instead of the full training data. Therefore, our task is to learn such partitioning approximately (since constraining the partitions to have exact equivalence of strategies is a hard problem). Since it is hard to know apriori how many partitions are needed, we formulate this as a non-parametric clustering problem and use DP-Means~\cite{kulisJ12} to learn the clusters. 

\begin{figure}
    \centering
    \includegraphics[scale=0.55]{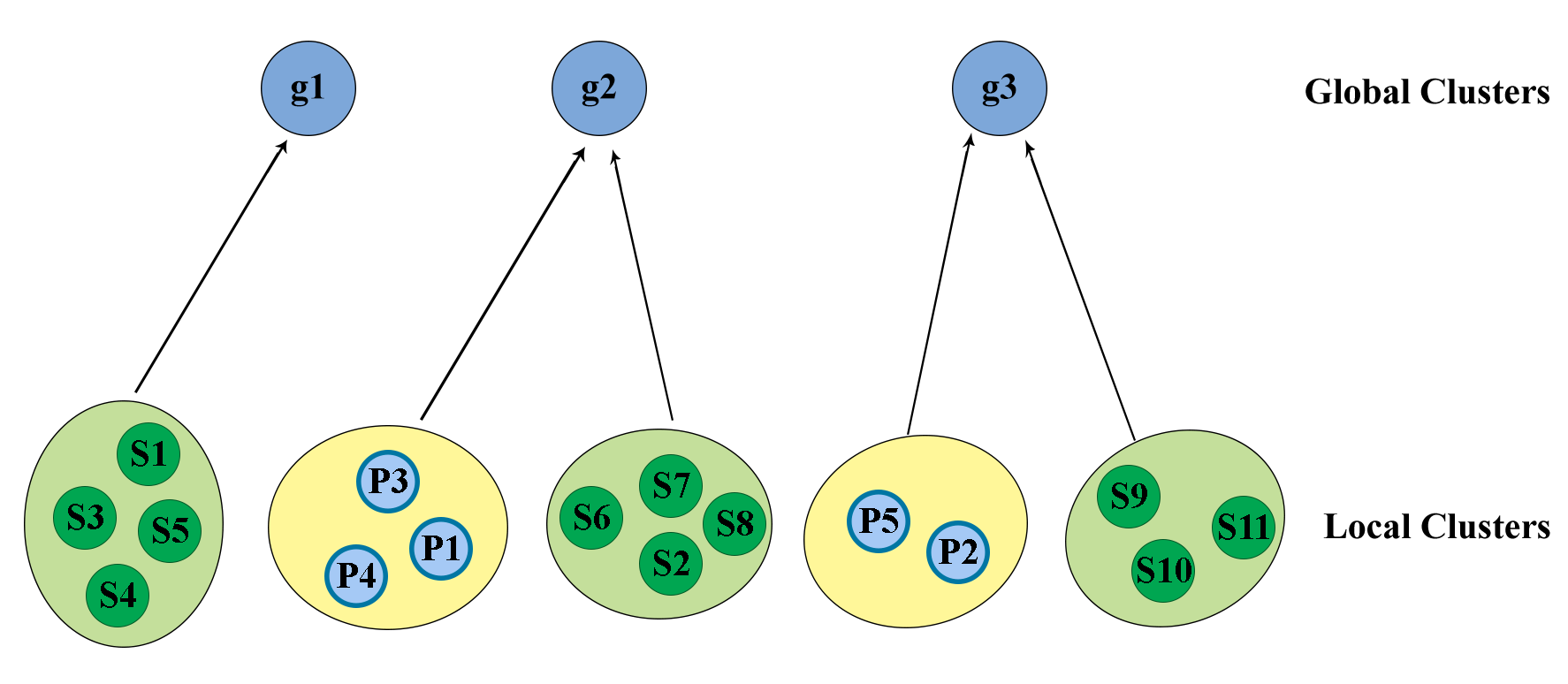}
    \caption{HDP Clustering showing the local clusters (student clusters and problem clusters) and the global clusters that combine the student, problem clusters.}
     \Description{Illustrating the formation of HDP clusters.}
    \label{fig:hdp-clusters}
\end{figure}

To formalize our approach, we begin with some notation. 
Let ${\bf S}$ $=$ $\{x_{i1}\}_{i=1}^N$ denote the set of students and ${\bf P}$ $=$ $\{x_{j2}\}_{j=1}^M$ denote the set of problems.
We refer to the student and problem clusters as the {\em local} clusters. A {\em global} cluster combines student and problem clusters as illustrated in Fig.~\ref{fig:hdp-clusters}. 
We run the standard DP-Means HDP clustering algorithm to optimize Eq.~\eqref{eq:dpobj} and learn global clusters that combine local clusters over ${\bf S}$ and ${\bf P}$. Note that large values of the global penalty $\lambda_g$ result in a coarse clustering with few clusters and small values of the penalty result in fine-grained clusters. We adaptively change $\lambda_g$ where we progressively lower the penalty  yielding a {\em coarse-to-fine} refinement of the clusters. Specifically, suppose $\ell_1\ldots\ell_g$ are the current global clusters, we compute a score $\mathcal{S}(\ell_1\ldots\ell_g)$ based on the symmetry of strategies within each cluster and as long as the score progressively improves across iterations, we reduce $\lambda_g$ to obtain finer-grained clusters.



\begin{algorithm}
    \label{alg:cfs}
    \caption{Coarse-to-Fine Refinement}
    \algsetup{
        linenodelimiter=.
    }
    \begin{algorithmic}[1]
        \renewcommand{\algorithmicrequire}{\textbf{Input:}}
        \renewcommand{\algorithmicensure}{\textbf{Output:}}
        \REQUIRE Student/Problem set: \{$x_{ij}$\}, Constant Penalty parameter: $\lambda_{\ell}$, iteration limit $T$
        \ENSURE  Global strategy clustering \{$\ell_1,\dots,\ell_g$\}\eat{ and number of clusters $g$}
        \\ \textit{Initialize} : Global cluster penalty $\lambda_g = y$ (where $y$ is a large number), $t=0$, cluster coherence $coh_{t-1} = 0$.
        \REPEAT
            \STATE $t = t+1$
            \STATE Cluster with penalties $\lambda_{\ell}$, $\lambda_{g}$
            \STATE Compute cluster coherence score, $coh_t$ = $\mathcal{S}(\ell_1,\ldots\ell_g)$
            \STATE Reduce: $\lambda_{g} = \lambda_{g} - \epsilon$
        \UNTIL{$coh_{t} > coh_{t-1} $ or $t > T$}
    \end{algorithmic}
    \label{alg:cfs}
\end{algorithm}

\subsection{Refining Clusters using Symmetry}
\label{section:strategy_symmetry}

Note that each global cluster implicitly represents a set of strategies, i.e., a student-problem pair $(s,p)$ within the cluster corresponds to a strategy followed by $s$ for problem $p$. We want to quantify symmetry between strategies within a cluster. A naive approach to compare two strategies is to compute the mean of the MVec embeddings for the KCs used in each strategy and then compute the distance between the means.
However, this  assumes that all permutations of a strategy are equivalent to each other which is problematic. On the other hand, suppose we compare the KC embedding at a step in one strategy with the KC embedding at the same step in the other strategy, then we assume the strategies are equivalent only if they are perfectly aligned with each other which is also an over-simplification.
To match strategies approximately, we represent a strategy using a combination of embeddings and positional encodings~\cite{vaswani&al17}, and approximately align two strategies to estimate the symmetry between them.
A KC $K$ in the strategy is represented by its positional embedding $\vec{K}$ $=$ $\vec{K_e}$ $+$ $\vec{K_p}$ where $\vec{K_e}$ is the MVec embedding for $K$ and $\vec{K_p}$ is the positional encoding for $K$ in the strategy. To compute symmetry between strategies, we compute an alignment between their positional embeddings.
Alignment is a fundamental problem in domains such as Bioinformatics where a classical approach that is often used is the Smith and Waterman algorithm (SW)~\cite{smith_waterman}. The idea is to perform local search to compute the best possible alignment between two sequences. SW requires a {\em similarity function} which in our case is the similarity between two KCs and we set this to be $s(K,K')$ $=$ $\vec{K}^\top\vec{K}'$, i.e., the cosine similarity between the positional embeddings of $K$ and $K'$. Further, SW also requires a {\em gap penalty} which refers to the cost of leaving a gap in the alignment. In our case, we set the gap penalty to 0 since we want symmetry between strategies to be invariant to gaps. That is, if two strategies are symmetric, adding extra steps in the strategies is acceptable. SW iteratively computes a scoring matrix based on local alignments. The worst-case complexity to compute the scoring matrix that gives us scores for the best alignment is equal to $O(m*n)$ where $m$ and $n$ are lengths of the strategies. 

Note that in our case, we are interested in quantifying symmetry between strategies based on the alignment. Specifically, let ${\bf K}$ and ${\bf K}'$ be two strategies of lengths $n$ and $m$ respectively. SW gives us an alignment between ${\bf K}$ and ${\bf K}'$ denoted by $L({\bf K},{\bf K}')$. The alignment consists of the pairs KCs from ${\bf K}$ and ${\bf K}'$ respectively that have been matched/aligned or a gap, i.e., a KC from ${\bf K}$ could not be aligned with any KC from ${\bf K}'$. We compute the symmetry score between ${\bf K}$ and ${\bf K}'$ as $r({\bf K},{\bf K}')$ $=$ $\frac{1}{\max(n,m)}$ $\sum_{(K,K')\in L({\bf K},{\bf K}')}$ $(\vec{K}^\top\vec{K'})$, where $(K,K')$ $\in$ $L({\bf K},{\bf K}')$ are aligned KCs and $\vec{K}^\top\vec{K'}$ is their cosine similarity. We see that $0\leq$ $r({\bf K},{\bf K}')$ $\leq 1$. Based on this, we estimate symmetry in the clustering as follows.
\eat{
\begin{align}
\label{eq:sym}
    \mathcal{S}(\ell_1,\ldots,\ell_g)=
    &\frac{1}{g}\sum_{p=1}^g\frac{2}{|T(\ell_p)| (|T(\ell_p)|-1)}\\\nonumber
    &\sum_{{\bf K},{\bf K'}\in T(\ell_p)}r({\bf K},{\bf K}')
\end{align}
}
\begin{align}
\label{eq:sym}
    \mathcal{S}(\ell_1,\ldots,\ell_g)=
    &\frac{1}{g}\sum_{p=1}^g \mathbb{Z}_p \sum_{{\bf K},{\bf K'}\in T(\ell_p)}r({\bf K},{\bf K}')
\end{align}
$T(\ell_p)$ is a set of all strategies in $\ell_p$ and $\mathbb{Z}_p=\frac{2}{|T(\ell_p)|(|T(\ell_p)|-1)}$ is the normalization term. Thus, a larger value of $\mathcal{S}(\ell_1,\ldots,\ell_g)$ implies that the clustering corresponding to $\ell_1,\ldots,\ell_g$ has a greater degree of symmetry in strategies. Using this score, we refine the clustering by adapting the global penalty. Specifically, we reduce the global penalty $\lambda_g$ by a constant $\epsilon$ as long as the symmetry score decreases across iterations or for a fixed number of iterations. Algorithm~\ref{alg:cfs} summarizes the coarse-to-fine refinement.

\subsection{Training the Model}
We use an LSTM architecture similar to~\cite{shakya2021} to predict strategies. Specifically, the model is a one-to-many LSTM that takes student, problem vectors as input and generates a sequence of KCs as output. To train this model, we sample instances from the converged global clusters, and for each sampled student-problem pair $(s,p)$, the LSTM input is the concatenation of MVec embeddings of $s$ and $p$. The output corresponds to the sequence of KCs in the strategy used by $s$ for $p$, each of which is encoded as a one-hot vector. To handle variable-length strategies, a special {\em stop} symbol is used to denote the end of a sequence. The entire model is trained using the standard categorical cross-entropy loss.

\section{Experiments}

Our goal is to answer the following questions through our evaluation. i) what is the accuracy of our approach in predicting strategies? ii) how does our approach scale-up? iii) what is the influence of mastery in predicting strategies accurately? and iv) is there a disparity in the accuracy of prediction for different skill-based sub-groups in the data?

\subsection{Dataset}
The data we use in this work is large-scale real-world education data recorded with real students using MATHia. MATHia is an online math learning program for middle school students that is popularly used across several schools. We used two datasets provided by MATHia for evaluating our proposed approach, Bridge-to-Algebra 2008-09 (\texttt{BA08}) and Carnegie Learning MATHia 2019-20 (\texttt{CL19}). Both of these datasets contain recorded interactions between the student and the computer tutor while the student attempts to solve a problem on the platform. Each recorded interaction consists of the log of the student's action toward solving the problem, for example, the knowledge component used, if hints were needed and if the step was completed correctly. \texttt{BA08} is an older dataset that consists about $20$ million interactions for about $6000$ students and $52k$ unique algebra problems. This dataset contains about $1.6$ million \textit{data instances}. It is important to note that we consider a \textit{data instance} as a student-problem pair, so one \textit{data instance} consolidates all the interactions/steps for one student on a specific problem. \texttt{CL19} is a more recent and larger dataset containing about $47$ million interactions for $5000$ students and about $32k$ unique math problems. It has about $1.9$ million \textit{data instances}. Both datasets are publicly available through the PSLC datashop~\cite{StamperKBSLDYS11a}.


\subsection{Experimental Setup}
To train the attention model, we used the transformer implementation in~\cite{vaswani&al17}. For the strategy prediction, we used a one-to-many LSTM~\cite{shakya2021} where the input is the student and problem embedding, and the output is the sequence of KCs. The parameters for the two models are shown in Table \ref{tab:params}. We used the standard parameters for the transformer model and retained the same parameters as in ~\cite{shakya2021} for the LSTM model for an unbiased comparison. For generating MVec embeddings, we used Gensim~\cite{mikolov&al13} with an embedding dimension set to 300 (which is typically used). We initialize the local cluster penalty $\lambda_{\ell} = 7$ and global cluster penalty $\lambda_g = 9$ for Coarse-to-Fine refinement and reduce the global penalty by $\epsilon$ $=$ 1 (we discovered these to be the best-performing hyper-parameters in experimentation). We perform our experiments on a machine with 64 GB RAM, an Nvidia Quadro 5000 GPU with 16 GB memory, and a CPU with 8 cores. The code for our implementation is available here
\footnote{\url{https://github.com/anupshakya07/attn-scaling}}.

\begin{table}[!t]
    \centering
    \caption{Main parameters for the models.}
    \label{tab:params}
    \resizebox{0.48\textwidth}{!}{
    \begin{tabular}{c||c}
        \hline
        {\bf Transformer-based model} & {\bf LSTM-based strategy model}\\
        \hline\hline
        Dimension $\rightarrow$ 512 & Latent Dimension $\rightarrow$ 200\\
        Number of layers $\rightarrow$ 6 & Epochs $\rightarrow$ 60\\
        Number of heads $\rightarrow$ 8 & Batch Size $\rightarrow$ 30\\
        Dimensions of key, value and query $\rightarrow$ 64 & Adam Optimizer with Learning rate 0.01\\
        Max Sequence Length $\rightarrow$ 150  & Dropout $\rightarrow$ 0.1\\
        Dropout $\rightarrow$ 0.1 &\\
        Weight Sharing $\rightarrow$ False &\\
        \hline
    \end{tabular}
    }
\end{table}


\subsection{Comparison to Baselines}
We compared our approach with the following methods. The first one is a specialized approach proposed in Shakya et. al.~\cite{shakya2021} ($\mathrm{CS}$) for the same datasets where an LSTM is trained using importance sampling. However, this sampling does not incorporate mastery or approximate symmetries to find diverse training instances.
We also applied a more general importance sampling approach that is said to be applicable for any DNN model training proposed in~\cite{KatharopoulosF19} ($\mathrm{IS}$) using their publicly available implementation. However, $\mathrm{IS}$ failed to output any results for datasets of our size and therefore we do not show it in our result graphs. This indicates that general-purpose methods do not scale up for our datasets. We also developed a stratified sampler ($\mathrm{GS}$ for group sampling) where the distribution is only proportional to the number of problems solved by a student, i.e., we sample more instances from students that have data associated with them. The last baseline is a naive Random Sampler ($\mathrm{RS}$) used as a validation check where we sample students and problems uniformly at random. We refer to our approach as Attention Sampling ($\mathrm{AS}$).
In our evaluation, for each approach, we enforce a limit on the number of training instances and measure test accuracy based on the model trained with this limit. This is similar to a measure of the {\em effective model complexity}~\cite{nakkiranKBYBS20} which is the number of training samples to achieve close to zero error. We report the average accuracy of predicted KCs based on three training runs.


\begin{figure*}
    \centering
    \begin{subfigure}[t]{0.33\textwidth}
        \centering
        \includegraphics[width=\textwidth]{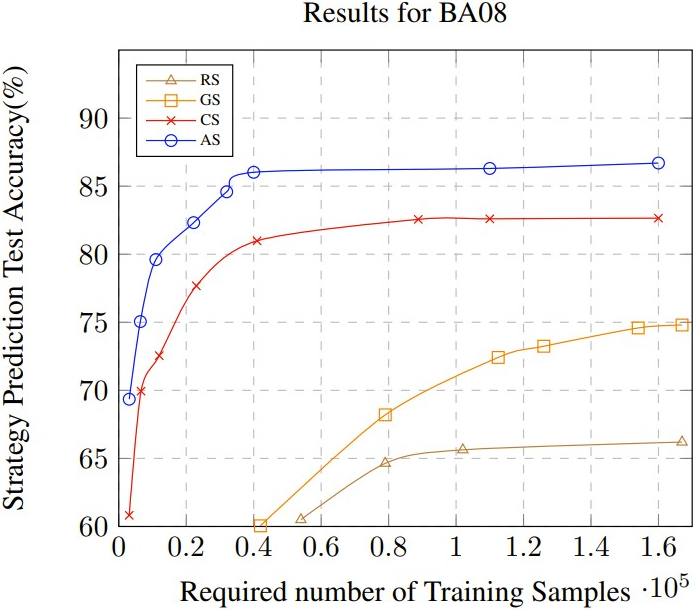}
        \Description{Results of Accuracy vs. Num. of Samples for BA08}
    \end{subfigure}
    \begin{subfigure}[t]{0.33\textwidth}
        \centering
        \includegraphics[width=\textwidth]{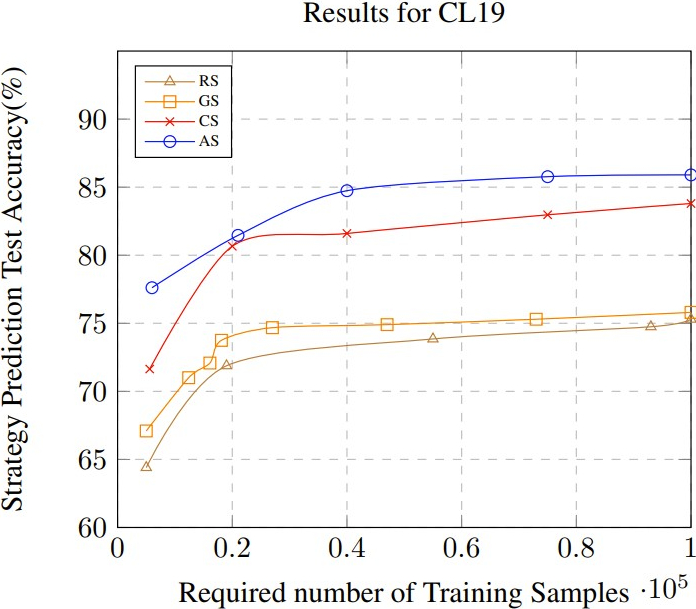}
        \Description{Results of Accuracy vs. Num. of Samples for CL19}
    \end{subfigure}
    \begin{subfigure}[t]{0.33\textwidth}
        \centering
        \includegraphics[width=\textwidth]{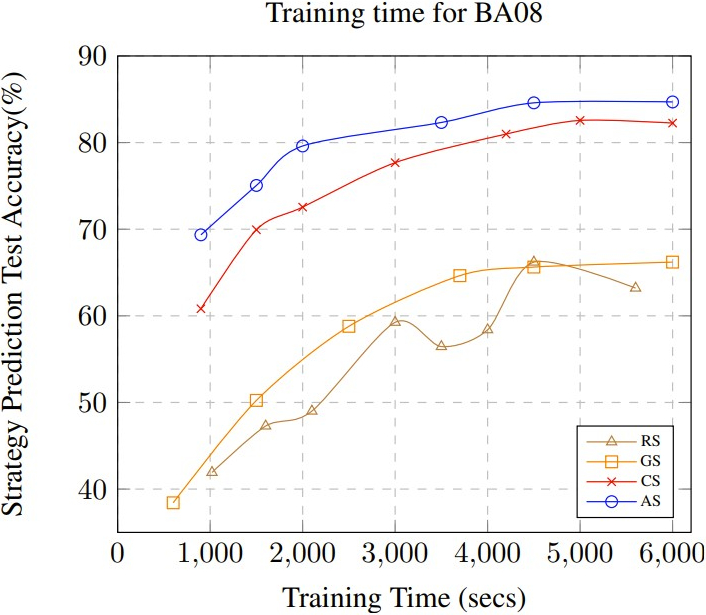}
        \Description{Results of Accuracy vs. Training Time for BA08}
    \end{subfigure}
    \begin{subfigure}[t]{0.33\textwidth}
        \centering
        \includegraphics[width=\textwidth]{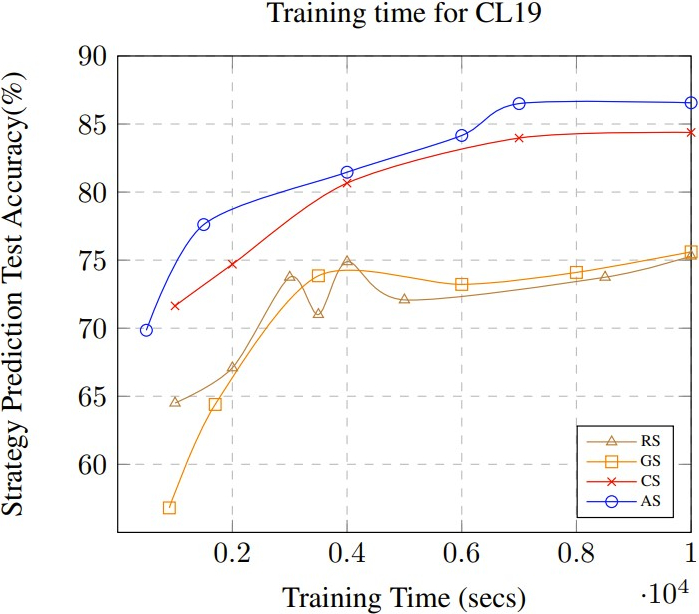}
        \Description{Results of Accuracy vs. Training Time for CL19}
    \end{subfigure}
    \caption{Illustrating Scalability vs Accuracy. (a), (b) show test accuracy for strategy prediction for varying training datasize limits.  (c), (d) show accuracy (strategy prediction) for different training time limits.}
    \label{fig:stratpred}
\end{figure*}

\subsection{Results and Discussion}

\subsubsection{Accuracy}
The strategy prediction accuracy results for \texttt{BA08} and \texttt{CL19} are shown in Fig.~\ref{fig:stratpred} (a) and (b). As shown in Fig.~\ref{fig:stratpred} (a), for \texttt{BA08}, in our approach ($\mathrm{AS}$), it takes less than $1\%$ of the entire data (of \texttt{BA08} containing 1.6 million instances) to obtain test accuracy that is greater than 80$\%$. $\mathrm{CS}$ is the next best performer but is consistently below $\mathrm{AS}$ for all training sizes. $\mathrm{GS}$ performs significantly worse which illustrates that symmetries are more complex and a simple grouping based on problems/students is insufficient. The poor performance of $\mathrm{RS}$ validates that the problem of choosing the correct samples is a challenging one. As seen in Fig.~\ref{fig:stratpred} (b), for a considerably larger dataset \texttt{CL19}, we can observe similar performance as in \texttt{BA08}. AS remains the best performer and here $\mathrm{CS}$ is less stable since we see a performance drop as we increase the limit on training samples. This suggests that $\mathrm{CS}$ may not be able to capture all symmetries and thus may produce a more biased training sample set. The results for $\mathrm{GS}$ and $\mathrm{RS}$ are similar to those observed in \texttt{BA08}. As mentioned before, $\mathrm{IS}$ failed to produce any results.

\subsubsection{Scalability} Fig.~\ref{fig:stratpred} (c) and (d) show the training time required to obtain a specific accuracy for \texttt{BA08} and \texttt{CL19} respectively. Even with the extra processing that is needed to compute the mastery-based embeddings and the non-parametric clustering, $\mathrm{AS}$ requires the shortest training time to achieve an accuracy that is higher than the other approaches. This illustrates the significance of leveraging symmetries in the data to train the model. As mentioned before, the full data is infeasible to train and when attempting to use the full data, the model did not converge for both datasets even after several days of training time using our experimental setup. As seen in our results, for \texttt{CL19}, the training time is larger since it takes longer to compute the groups using non-parametric clustering due to the much larger size of the dataset. However, considering that \texttt{CL19} is significantly larger than \texttt{BA08}, we see that AS could still scale up to this dataset quite easily while IS which is a state-of-the-art sampling method for DNN training failed to train the model.

\begin{table}[]
    \centering
    \caption{Ablation study with NS (No symmetries used), SS (Symmetries without using mastery) and MS (Adding the mastery model to better identify symmetries). Results are shown for 2 datasets with different sample sizes. Accuracy results in \%.}
    \scalebox{0.99}{
        \begin{tabular}{|c|c|c|c|c|c|c|}
            \hline
             \multirow{2}{*}{
             \textbf{Expts.}} & \multicolumn{3}{c|}{\textbf{BA08}} & \multicolumn{3}{c|}{\textbf{CL19}}  \\
             \cline{2-7}
             & \textbf{40k} & \textbf{100k} & \textbf{150k} & \textbf{40k} & \textbf{80k} & \textbf{100k} \\
             \hline
             \textbf{NS} & 60.05 & 71.14 & 74.58 & 74.81 & 75.4 & 75.8\\
             \hline
             \textbf{SS} & 80.98 & 82.3 & 82.65 & 81.6 & 83.2 & 83.8\\
             \hline
             \textbf{SS + MS} & {\color{Green} 86.02} & {\color{Green} 86.21} & {\color{Green} 86.53} & {\color{Green} 84.74} & {\color{Green} 85.8} & {\color{Green} 85.9}\\
             \hline
        \end{tabular}
    }
    
    \label{tab:ablation}
\end{table}

\begin{figure*}
    \centering
    \begin{subfigure}[t]{0.35\textwidth}
        \centering
        \includegraphics[width=\textwidth]{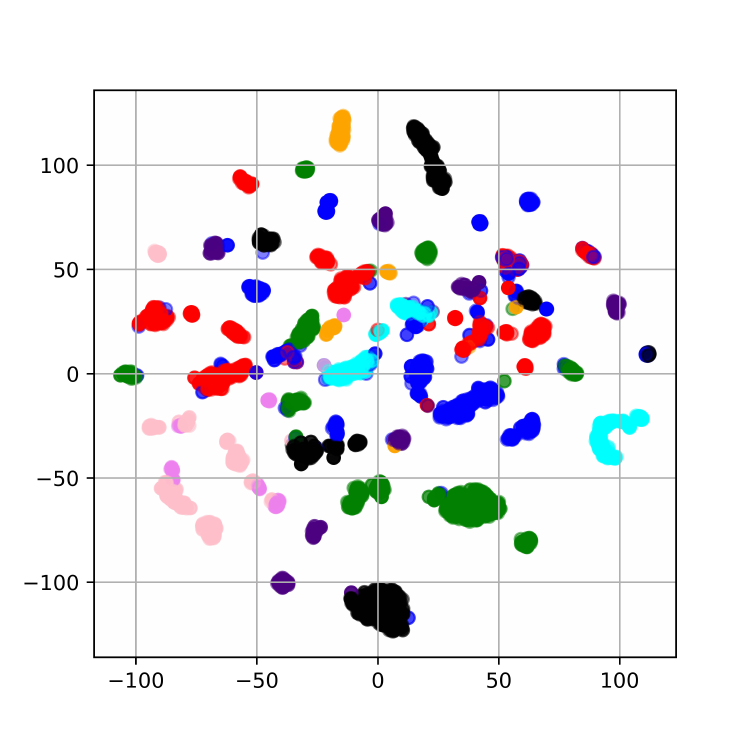}
        \Description{Different analysis on the results. TSNE-plot for simple Word2Vec}
    \end{subfigure}
    \begin{subfigure}[t]{0.35\textwidth}
        \centering
        \includegraphics[width=\textwidth]{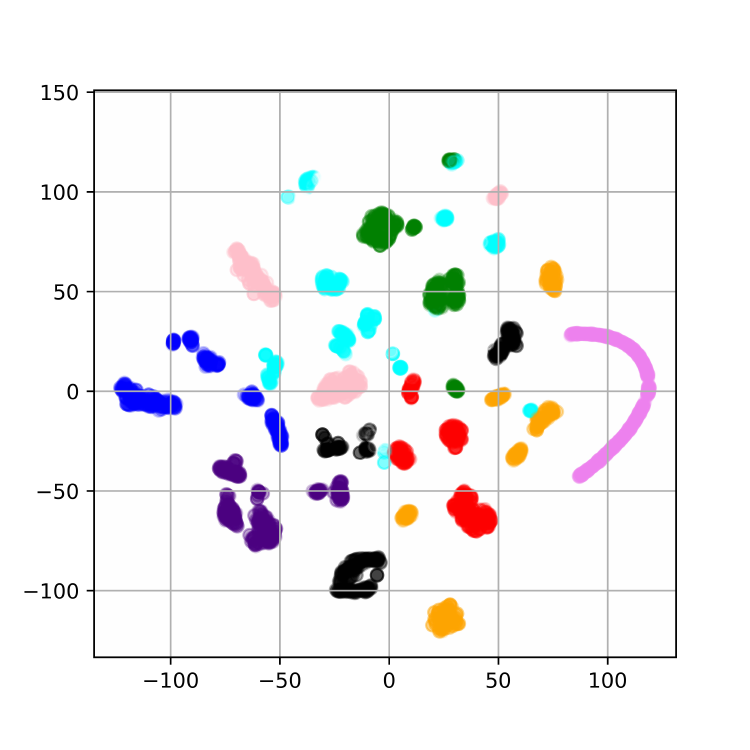}
        \Description{Different analysis on the results. TSNE-plot for Mvec Embeddings.}
    \end{subfigure}
    \begin{subfigure}[t]{0.35\textwidth}
        \centering
        \includegraphics[width=\textwidth]{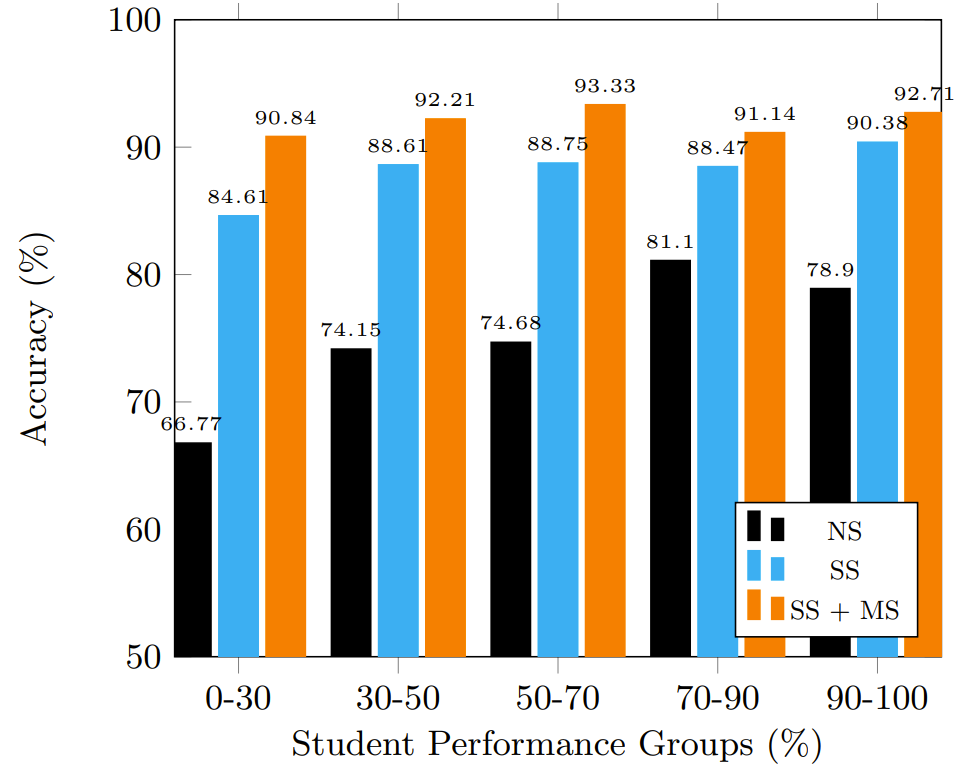}
        \Description{Different analysis on the results. Fairness Plot.}
    \end{subfigure}
    \begin{subfigure}[t]{0.35\textwidth}
        \centering
        \includegraphics[width=\textwidth]{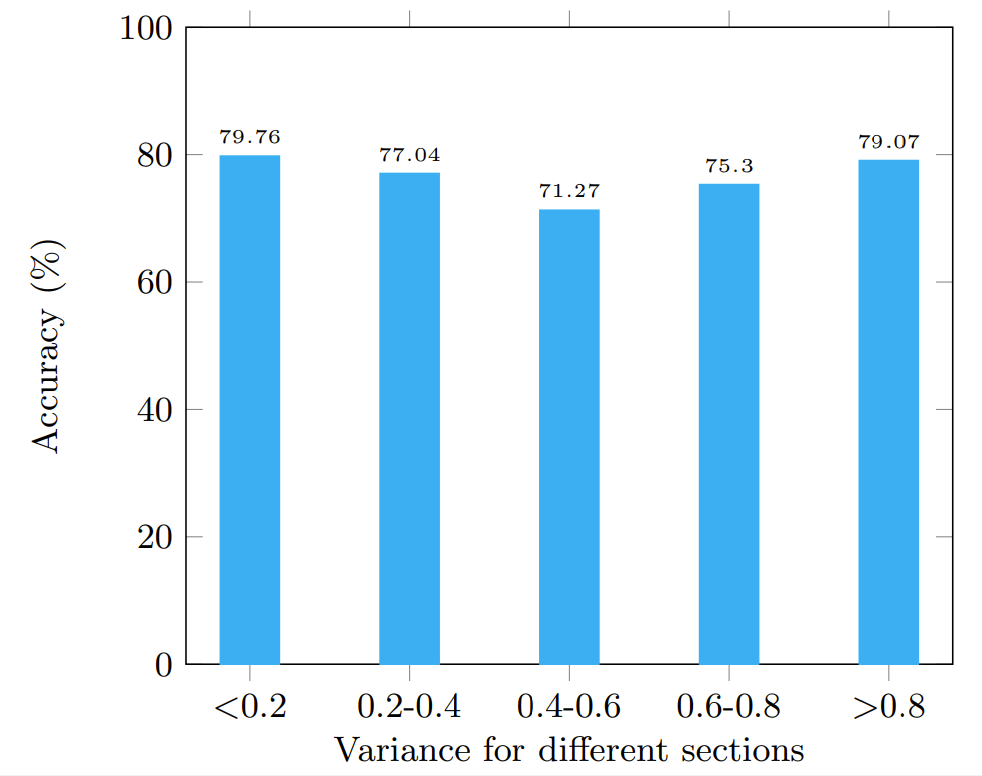}
        \Description{Different analysis on the results. Section-wise Variance.}
    \end{subfigure}

    \caption{T-SNE visualization of strategy clusters for \texttt{CL19}. The color-coded plots show the 2D representation of the different strategy clusters for (a) Embeddings that do not use mastery (b) MVec embeddings. The strategy representations are extracted from the final hidden layer of the LSTM model and converted to 2D representation using T-SNE. (c) shows accuracy for different groups of students (based on their performance) for \texttt{CL19}. The x-axis denotes different ranges of \%s, where a range $a-b$ denotes that students in this group got $>a$ and $<b$ steps correct in their first attempt. The y-axis shows accuracy over the groups. \eat{As seen here, our model was fair where the accuracy is invariant to performance-based groups and outperforms the model without mastery. }(d) shows the performance of the model on different groups based on the average variance of the strategies in the sections for \texttt{CL19}. Variance is computed using edit distance as the metric of similarity between strategies.}
    \label{fig:strat-invariance}
\end{figure*}
\begin{figure*}
    \centering
    \includegraphics[scale=0.79]{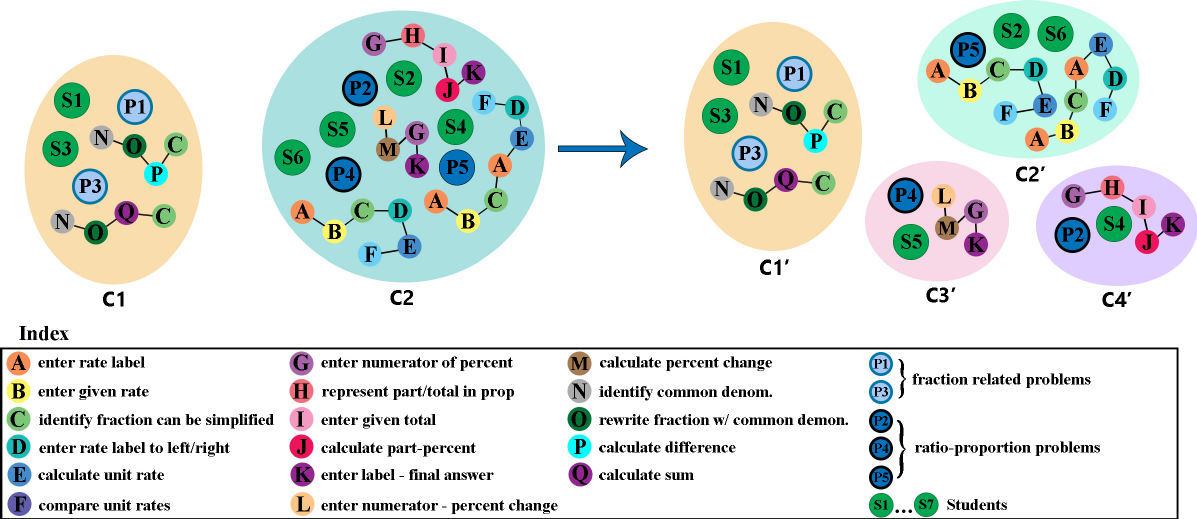}
    \caption{An example from the dataset CL19 illustrating coarse-to-fine refinement of clusters. Strategies are shown by paths connecting KCs. ${\bf C1}$ and ${\bf C2}$ are the coarse clusters which get refined into strategy invariant clusters ${\bf C1}'$, ${\bf C2}'$, ${\bf C3}'$ and ${\bf C4}'$.}
    \Description{Illustrating Coarse-to-fine refinement.}
    \label{fig:mvec_clustering}
\end{figure*}

\begin{table*}[]
    \caption{\label{tab:strategy-switch}Different strategies used by the students for different problems in the same section for \texttt{CL19} dataset. The model is able to predict accurately as student adapt their strategies.}
    \resizebox{1\textwidth}{!}{
    \begin{tabular}{|c|p{8em}|l|l|}
        \hline
        \cellcolor{gray!25} \textbf{Student} & \cellcolor{gray!25} \textbf{Problem Name} & \cellcolor{gray!25}{\bf Predicted Strategy} & \cellcolor{gray!25}{\bf Actual Strategy}\\
        \hline
        \cellcolor{blue!25} & \multirow{7}{0.15\textwidth}{linear inequalities numberline 5}  & represent open point on numberline-1 & represent open point on numberline-1\\
        \cellcolor{blue!25} &  & represent ray on numberline-1 & represent ray on numberline-1\\
        \cellcolor{blue!25} &  & represent inequality in symbolic problem-1 & represent inequality in symbolic problem-1\\
        \cellcolor{blue!25} &  & identify when finished with numberline-1 & identify when finished with numberline-1\\
        \cellcolor{blue!25} &  & identify invisible non-inflection point is in solutionset-1 & identify invisible non-inflection point is in solutionset-1\\
        \cellcolor{blue!25} &  & identify invisible non-inflection point is not in solutionset-1 & identify invisible non-inflection point is not in solutionset-1\\
        \cellcolor{blue!25} &  & identify visible non-inflection point is not in solutionset-1 & identify visible non-inflection point is not in solutionset-1\\
        \cline{2-4}
        \cellcolor{blue!25} & \multirow{9}{0.15\textwidth}{linear inequalities numberline 9} & write simple inequality in verbal problem-1 & write simple inequality in verbal problem-1\\
        \cellcolor{blue!25} &  & represent closedpoint on numberline-1 & represent closedpoint on numberline-1\\
        \cellcolor{blue!25} &  & represent ray on numberline-1 & represent ray on numberline-1\\
        \cellcolor{blue!25} &  & identify when finished with numberline-1 & identify when finished with numberline-1\\
        \cellcolor{blue!25} &  & identify visible non-inflection point is not in solution set-1 &  identify visible non-inflection point is not in solution set-1\\
        \cellcolor{blue!25} &  & identify invisible non-inflection point is not in solution set-1 &  identify invisible non-inflection point is not in solution set-1\\
        \cellcolor{blue!25} &  & identify inflection point in solution set-1 &  identify inflection point in solution set-1\\
        \cellcolor{blue!25} &  &\cellcolor{red!10}  & \cellcolor{red!10}  identify invisible non-inflection point is not in solution set-1\\
        \multirow{-16}{*}{\cellcolor{blue!25} $S_1$} & & \cellcolor{red!10}  &\cellcolor{red!10} identify inflection point in solution set-1\\
        \hline

        \cellcolor{yellow!25} & \multirow{10}{0.15\textwidth}{ratio proportion prop1 4} & enter part in proportion with variable-1 & enter part in proportion with variable-1\\
        \cellcolor{yellow!25} & & enter given total in proportion-1 & enter given total in proportion-1\\
        \cellcolor{yellow!25} & &\cellcolor{red!10} enter numerator of given rate in proportion-1 & \cellcolor{red!10} enter denominator of given rate in proportion-1\\
        \cellcolor{yellow!25} & &\cellcolor{red!10} enter denominator of given rate in proportion-1 & \cellcolor{red!10} enter numerator of given rate in proportion-1\\
        \cellcolor{yellow!25} & & enter proportion label in numerator-1 & enter proportion label in numerator-1\\
        \cellcolor{yellow!25} & & enter proportion label in denominator-1 & enter proportion label in denominator-1\\
        \cellcolor{yellow!25} & & calculate part in proportion with fractions-1 & calculate part in proportion with fractions-1\\
        \cellcolor{yellow!25} & &\cellcolor{red!10} enter numerator of form of 1-1 & \cellcolor{red!10} enter denominator of form of 1-1\\
        \cellcolor{yellow!25} & &\cellcolor{red!10} enter denominator of form of 1-1 &\cellcolor{red!10} enter numerator of form of 1-1\\
        \cellcolor{yellow!25} & & enter calculated value of rate-1 & enter calculated value of rate-1\\
        \cline{2-4}
        \cellcolor{yellow!25} & \multirow{7}{0.15\textwidth}{ratio proportion prop1 5} & enter proportion label in numerator-1 & enter proportion label in numerator-1\\
        \cellcolor{yellow!25} & & enter proportion label in denominator-1 & enter proportion label in denominator-1\\
        \cellcolor{yellow!25} & & enter given total in proportion-1 & enter given total in proportion-1\\
        \cellcolor{yellow!25} & & enter numerator of given unit rate in proportion-1 & enter numerator of given unit rate in proportion-1\\
        \cellcolor{yellow!25} & & enter denominator of given unit rate in proportion-1 & enter denominator of given unit rate in proportion-1\\
        \cellcolor{yellow!25} & & calculate part in proportion with fractions-1 & calculate part in proportion with fractions-1\\
        \multirow{-17}{*}{\cellcolor{yellow!25} $S_2$}& & enter calculated value of rate-1 & enter calculated value of rate-1\\
        \hline
        
    \end{tabular}
    }
    
\end{table*}

\subsubsection{Ablation Study} 

Table~\ref{tab:ablation} shows the results of our ablation study. We add each component to our overall approach and observe the test accuracy as we vary the sample size in the training data. Specifically, the first case ({\bf NS}) uses no symmetries, i.e., the clustering is performed randomly. Next, we cluster based on embeddings without using the mastery, i.e., when we generate the embeddings for MVec, we do not use the attention model and simply use triplets $(S,P,K)$, where $S$ is a student, $P$ is a problem and $K$ is a KC used by $S$ for $P$ as input to Word2Vec and generate embeddings. Thus, we use symmetries in strategy without utilizing mastery when we generate the clusters. We show this as Strategy Symmetry ({\bf SS}) in the table. Finally, we add mastery to generate embeddings denoted by {\bf SS} + {\bf MS} and as shown, this improves the generalization performance for all sample-sizes thus, illustrating that utilizing mastery to learn embeddings plays a significant role in improving accuracy in predicting strategies.


\subsubsection{Visualizing Clusters}
We used T-SNE  to visualize the clusters of strategies. For this, we pick 100 student-problem pairs sampled from 10 clusters. We then perform strategy prediction for these and visualize the hidden-layer representation of the LSTM in the T-SNE plot. We compare this for MVec embeddings as well as embeddings that are learned without using mastery. As shown in Fig. \ref{fig:strat-invariance} (a) and (b), when we use MVec, the LSTM hidden-layer representation of strategies has better separation. This indicates that we learn better grouping of strategies using MVec embeddings.

\subsubsection{Fairness} 
We evaluate if our approach results in disparate mistreatment. Specifically, this means that the model should not have significantly different accuracy for different sensitive sub-groups in the data. In our case, the sensitive sub-groups correspond to students at different skill levels. That is, we want to predict the strategies equally well for all students. To do this, we conducted an experiment where we divide the test data into 6 performance groups. The performance groups are based on the $\%$ of problem steps the students solve correctly on their first attempt. The performance groups include students who scored in the following ranges $\leq 30\%$, $30-50\%$, $50-70\%$, $70-90\%$, $\geq 90\%$. To measure disparate mistreatment, we compare the average accuracy of strategies predicted for each of these groups. For a student $S$ in performance group ${\bf G}$, we predict the strategies for all problems attempted by $S$ in the test set and measure the average accuracy $\mu_S$. We then compute the accuracy over a performance group as $1/|{\bf G}|\sum_{S\in{\bf G}}\mu_S$. Fig.~\ref{fig:strat-invariance}(c) shows our results for the variants, NS, SS and SS+MS (identical to those used in the ablation study) for CL19 (we show results on this since this is the larger and more recent dataset). As seen from our results, SS+MS yields the best accuracy over each performance group. Further, the accuracy over the poorest and the best performers is comparable to each other and not significantly different. Thus, there is no disparate mistreatment of any performance group shown by our approach.

Next, we want to verify if there is disparate mistreatment when we consider sub-groups that have rare strategies. To measure this, we divided the problem sections in the test set into groups based on the variance among strategies for problems in those sections. Specifically, to perform worst-case analysis, we used the edit distance to measure the variance of strategies within problems in a section. That is, if a pair of problems vary in two out of 10 steps, the edit distance is 0.2. We computed the variance in edit distances over all the problems in a section. We then obtained the sub-groups at 5 different thresholds of variance. Thus, groups that have large variance include more rare strategies, while groups that have smaller variance have fewer rare strategies. For all the problems in each of these sub-groups, we computed the average accuracy in strategy prediction. Fig.\ref{fig:strat-invariance}(d) shows our accuracy results over all the sub-groups. As seen here, we have no disparate mistreatment for any of the sub-groups. Thus, we show that even in cases where rare strategies are used by students, our approach predicts strategies with an accuracy that is very similar to cases where common strategies are used.


\subsubsection{Example Cases} 
Table~\ref{tab:strategy-switch} illustrates examples corresponding to two different students where we predict strategies for two problems taken from the same section in each case. Note that the students make modifications to their strategy to suit the problem context as seen in the examples, though the overall strategies are similar since the problems are from the same section. The model is able to successfully adapt and predict these strategy changes quite accurately. In the case of student $S_1$, for the second problem, the model predicted most of the steps except that the student had some redundant steps at the end which were not predicted by the model. In the case of $S_2$, for problem 1, the predicted strategy interchanged the order of a couple of steps that clearly does not significantly alter the underlying strategy.

We illustrate some examples of coarse-to-fine refinement in Fig.~\ref{fig:mvec_clustering}. Specifically, we show examples from two types of problems, {\em Fractions} and {\em Ratio, Proportions}. The clusters indicate the students, problems, and strategies followed by students. In cluster ${\bf C1}$, even though there are two different strategies, they are symmetric to each other and therefore, in a subsequent iteration of refinement, ${\bf C1}'$ is the same as ${\bf C1}$. On the other hand, ${\bf C2}$ consists of 4 strategies, 2 of these are expert-level strategies and the other two are simpler but differing strategies. Upon refinement of ${\bf C2}$, we get ${\bf C2}'$ which intuitively represents the expert students and ${\bf C3}'$, ${\bf C4}'$ which represents students using simpler yet different strategies. Thus, the coarse-to-fine refinement results in invariant strategies within each cluster.



\section{Conclusion}

We presented a scalable and equitable framework for predicting math problem-solving strategies used by students. Since students with differing skill levels use significantly different strategies, to predict these, we need to train a model over diverse training instances. Identifying such instances is a challenging problem in big data. Particularly, identifying strategies which are approximately symmetrical to each other is a hard task. Here, we developed a clustering approach to discover diverse groups where instances within each group have approximately symmetrical strategies. Specifically, we learned an embedding MVec using a combination of Node2Vec where we learned representations for relationships in the data encoded as a graph and a transformer model that predicts mastery. Specifically, similar attentions in the transformer model over steps in the strategy indicated similar mastery in solving a problem, which we used to learn the Node2Vec representation. We then clustered the MVec embeddings with a non-parametric algorithm called DP-Means by iteratively refining the clusters based on the level of symmetry encoded within the clusters.
By sampling from clusters, we were able to train an LSTM model to predict strategies using small but highly informative instances that were representative of strategies in the full data. Further, by sampling from clusters, we ensured that the LSTM model did not optimize its parameters for any specific group, but instead generalized over all groups in the data, thus making the model capable of identifying strategies from diverse groups. Experiments on two large-scale datasets demonstrated our accuracy in predicting strategies with a small fraction of the dataset and further, our predictions were fair across students at different levels of skill. 

As part of future work, we hope to extend this model to non-structured interactions (e.g. conversations). Further, we also plan to explore more complex mappings of strategies where each step can be represented by a structure (e.g. graph, table, etc.) and developing structured prediction models from such mappings. We also propose to utilize this approach in instructional design where we can select problems to solve based on a student's predicted strategy and also to develop interventions in ITSs based on misconceptions identified in predicted strategies.


\section{Acknowledgement} 
This research was sponsored by the National Science Foundation under NSF IIS award \#2008812 and NSF award \#1934745. The opinions, findings, and results are solely the authors' and do not reflect those of the funding agencies.
\bibliographystyle{abbrv}
\bibliography{main}

\end{document}